\documentclass[10pt,twocolumn,letterpaper]{article}

\usepackage[pagenumbers]{cvpr} %

\usepackage{bm}
\usepackage{stmaryrd}

\DeclareMathOperator*{\argmin}{arg\,min}
\usepackage[T1]{fontenc}

\makeatletter
\renewcommand{\paragraph}{%
  \@startsection{paragraph}{4}%
  {\z@}{1.5ex \@plus 1ex \@minus .2ex}{-1em}%
  {\normalfont\normalsize\bfseries}%
}
\makeatother

\usepackage[accsupp]{axessibility}

\usepackage{tablefootnote}
\usepackage{multirow}
\usepackage{colortbl}
\definecolor{lightgray}{gray}{0.9} 
\usepackage{arydshln}
\usepackage{stfloats}
\usepackage{booktabs}

\definecolor{cvprblue}{rgb}{0.21,0.49,0.74}
\usepackage[pagebackref,breaklinks,colorlinks,allcolors=cvprblue]{hyperref}

\def\paperID{16953} %
\def\confName{CVPR}
\def\confYear{2025}

\title{On-Device Self-Supervised Learning of Low-Latency\\Monocular Depth from Only Events}

\author{Jesse J. Hagenaars\,$^1$ \quad
Yilun Wu\,$^1$ \quad
Federico Paredes-Vall\'es\,$^2$ \\
Stein Stroobants\,$^1$ \quad
Guido C.H.E. de Croon\,$^1$ \\[0.5em]
$^1$ MAVLab, TU Delft \quad $^2$ EUISPC, Sony Semiconductor Solutions Europe, Sony Europe B.V. \\[0.5em]
\url{https://mavlab.tudelft.nl/depth_from_events}
}

\begin{document}
\twocolumn[{%
\renewcommand\twocolumn[1][]{#1}%
\maketitle
\vspace{-0.8cm}
\begin{center}
    \centering
    \captionsetup{type=figure}
    \includegraphics[width=\linewidth]{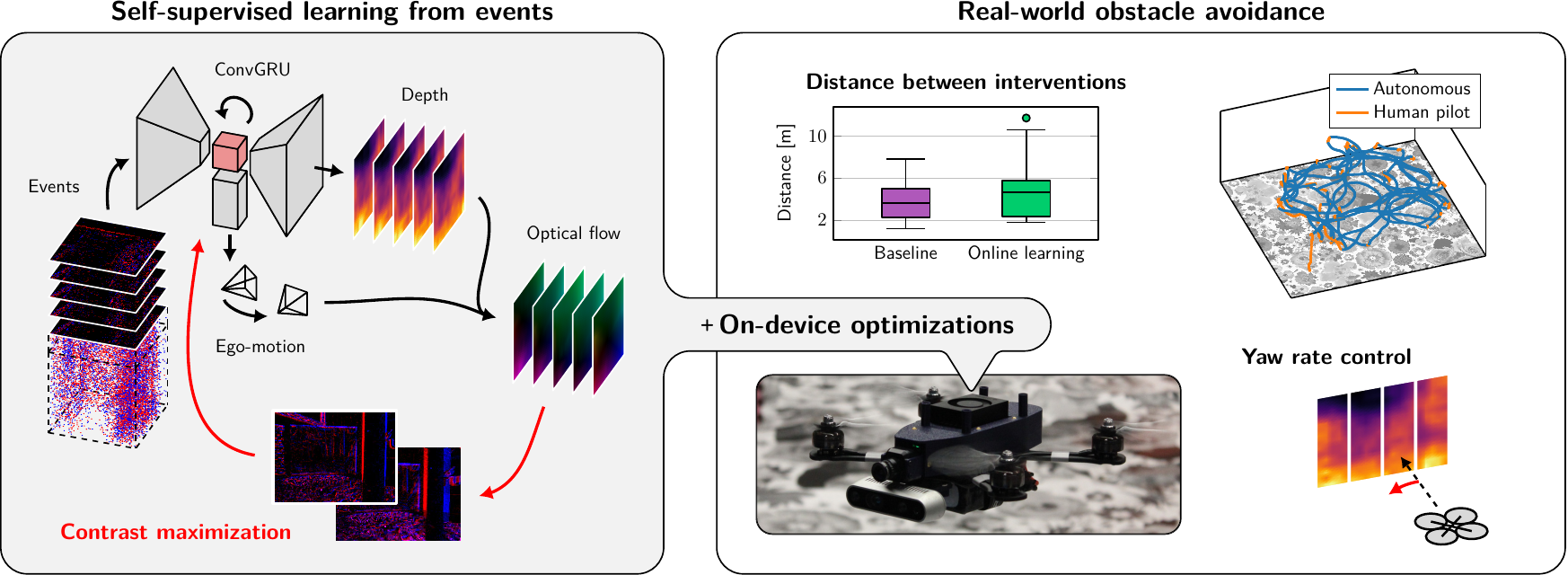}
    \captionof{figure}{Online, on-device learning allows robots to ``train in their test environment''. We improve the time and memory efficiency of the self-supervised contrast maximization pipeline, such that on-board learning of monocular depth from event camera data becomes possible. When deployed on a small drone, online learning leads to better depth estimates and more successful obstacle avoidance behavior.}
    \label{fig:overview}
\end{center}%
}]
\begin{abstract}

Event cameras provide low-latency perception for only milliwatts of power. This makes them highly suitable for resource-restricted, agile robots such as small flying drones. Self-supervised learning based on contrast maximization holds great potential for event-based robot vision, as it foregoes the need for high-frequency ground truth and allows for online learning in the robot's operational environment. However, online, on-board learning raises the major challenge of achieving sufficient computational efficiency for real-time learning, while maintaining competitive visual perception performance. In this work, we improve the time and memory efficiency of the contrast maximization pipeline, making on-device learning of low-latency monocular depth possible. We demonstrate that online learning on board a small drone yields more accurate depth estimates and more successful obstacle avoidance behavior compared to only pre-training. Benchmarking experiments show that the proposed pipeline is not only efficient, but also achieves state-of-the-art depth estimation performance among self-supervised approaches. Our work taps into the unused potential of online, on-device robot learning, promising smaller reality gaps and better performance.

\end{abstract}
    
\section{Introduction}
\label{sec:intro}

Event cameras capture per-pixel brightness changes at microsecond resolution, while consuming only milliwatts of power~\cite{gallego2020eventbased}. This combination enables low-latency perception and decision-making on agile but resource-constrained platforms such as small drones. %

To make full use of the temporal information in the event stream, the learning pipeline consisting of network architecture and loss function should also operate at high frequency~\cite{paredes-valles2023taming}. Ground truth optical flow or depth is often only available at lower rates of 10-20~Hz~\cite{zhu2018multivehiclea,gehrig2021dsec,chaney2023m3ed}. While some datasets allow upsampling of ground truth to higher rates~\cite{zhu2018multivehiclea,burner2022evimo2}, reaching the temporal resolution of an event camera might be difficult and come at the cost of high data rates. This holds back supervised learning at the short time scales perceivable with event cameras. 

Contrast maximization~\cite{gallego2018unifying} allows self-supervised learning (SSL) of optical flow, depth and ego-motion from events alone~\cite{zhu2019unsupervised,paredes-valles2023taming,shiba2024secrets,gallego2019focus}. Since no ground truth is needed, learning and prediction can run at higher frequencies of 100~Hz~\cite{paredes-valles2023taming} or even 200~Hz~\cite{paredes-valles2024fully}, with the main limiting factor being the network's ability to integrate information over increasingly sparse inputs as the rate increases. %

A major advantage of SSL is that it foregoes the costly process of obtaining ground truth, which enables learning to scale to large unlabeled datasets. An additional, but typically less emphasized advantage is that SSL can in principle be performed in the operational environment of a robot or other edge device. Such \emph{online} SSL greatly reduces the need for generalization of the learned model, as training happens directly on data sampled from the test distribution~\cite{vanhecke2018persistent}. For visual tasks like monocular depth estimation, this is particularly important, as generalization of this perceptual capability to environments different from the training environment is notoriously difficult~\cite{vodisch2023codeps,fu2024dectrain}. 

Online SSL, however,  introduces additional challenges. Chief among them is that not only the network but also the learning framework should be computationally efficient enough to run on board. In this work, we improve the efficiency of the contrast maximization pipeline such that on-device learning of low-latency monocular depth and ego-motion becomes feasible. We demonstrate continual learning of this complex visual task on board a small flying drone, and show the usability of the resulting depth for obstacle avoidance. Furthermore, we investigate various combinations of pre-training and on-board learning. When trained on event camera datasets, our small recurrent network shows state-of-the-art depth estimation performance among self-supervised approaches. Our work taps into the unused potential of on-board, online SSL, promising smaller reality gaps, leading to better performance.

\section{Related work}
\label{sec:related}

\paragraph{SSL of monocular depth.} Self-supervised learning of monocular depth has garnered significant attention since the early works that focused on joint depth-pose estimation for static scenes~\cite{zhou2017unsupervised, vijayanarasimhan2017sfmnet, godard2019digging}. These foundational studies have spurred further advancements to handle more complex scenarios, such as dynamic scenes, by integrating optical flow estimation~\cite{ranjan2019competitive, yin2018geonet}, leveraging regularization techniques~\cite{li2021unsuperviseda}, or incorporating motion segmentation~\cite{sun2023dynamodepth}. Additionally, several works have explored learning camera parameters~\cite{gordon2019depth, chen2019selfsuperviseda}, which is particularly relevant for on-device learning scenarios with unknown cameras.

Wang~\textit{et al.}~\cite{wang2019recurrent} demonstrate that recurrent networks can enhance depth estimation by effectively utilizing information from multiple frames, resulting in more consistent depth scale predictions. Similarly, Bian~\textit{et al.}~\cite{bian2019unsupervised} introduce a loss term to encourage scale consistency, addressing a critical challenge in depth estimation. Achieving depth predictions with a consistent scale significantly enhances the stability of robot control systems relying on these depth estimates. By combining recurrent architectures and scale-consistent training, our approach aligns with these advancements, offering a robust solution for on-device learning during real-world operation.

\paragraph{SSL through contrast maximization.} The contrast maximization framework enables the extraction of accurate optical flow information by leveraging the temporal misalignment of accumulated events~\cite{gallego2019focus, gallego2018unifying}. This optical flow can be estimated either through model-based methods~\cite{shiba2024secrets} or using neural networks~\cite{zhu2019unsupervised, hagenaars2021selfsupervised, paredes-valles2023taming}. The choice of the optical flow model itself offers a spectrum of possibilities, ranging from linear models~\cite{zhu2019unsupervised, hagenaars2021selfsupervised}, to segmented representations~\cite{paredes-valles2023taming}, and even parametrized trajectories~\cite{hamann2024motionprior}.

In our work, we adopt the approach outlined in~\cite{paredes-valles2023taming}, as it aligns with our goal of achieving high-frequency estimation using a neural network. Moreover, this approach supports efficient inference, which is critical for real-time applications such as robotic navigation or on-device learning. The ability to accommodate nonlinear event trajectories further broadens its applicability, making it a robust choice for extracting motion information in challenging conditions where traditional linear assumptions might fail.

\paragraph{SSL of depth from events.} Early works focused on jointly estimating depth and pose directly from events, employing either contrast maximization techniques~\cite{zhu2019unsupervised} or photometric error methods based on event frames~\cite{ye2020unsupervised}. A notable contribution by Zhu~\textit{et al.}~\cite{zhu2019unsupervised} was their ability to estimate metric depth through the incorporation of a stereo loss term, enabling absolute depth recovery. These methods demonstrated the potential of event-based sensing for depth and pose estimation in static scenes.

More recent research has expanded these approaches to address dynamic scenes~\cite{georgoulis2024out}, where traditional static-scene assumptions do not hold, and developed more principled frameworks for model-based contrast maximization to jointly estimate depth, ego-motion, and optical flow~\cite{shiba2024secrets}. These advancements represent a significant step toward estimating complex, real-world motion using event data.

Additionally, some works have explored the integration of intensity images as either inputs or components of the loss function~\cite{zhu2023selfsupervised}. This hybrid approach leverages the complementary information provided by intensity images to enhance the performance of event-based depth estimation, especially in scenarios where pure event data might lack sufficient structure or texture information.

\paragraph{Learning depth for drones.} Already in 2016, Lamers~\textit{et al.}~\cite{lamers2016selfsupervised} demonstrated the feasibility of learning depth estimation on board a small flapping-wing drone. Although their approach did not produce dense depth maps, it proved effective for navigation, marking an early milestone in on-device learning for aerial robotics. Several works~\cite{pirvu2021depth,licaret2022ufoa} combine unsupervised depth learning with an analytic odometry-flow pipeline (from external sources) to achieve metric monocular depth. While their networks are lightweight enough for embedded hardware, they are not actually used on drones.

Recent works have focused on generating dense depth maps on board drones. Liu~\textit{et al.}~\cite{liu2023nano} developed a system to estimate depth from images for obstacle avoidance on a tiny quadrotor, leveraging recorded real-world datasets for training. Bhattacharya~\textit{et al.}~\cite{bhattacharya2024monocular}, on the other hand, used a simulation-based approach to train depth estimation from events. They successfully transitioned to real-world applications by performing offline fine-tuning on real-world data, enabling effective obstacle avoidance in practice.

In this work, we take a distinct approach by performing fine-tuning on board the drone in an online fashion during flight, adapting the model in real time while actively avoiding obstacles. This method combines the strengths of self-supervised learning with real-time adaptability, paving the way for more robust and efficient systems capable of handling dynamic environments. By eliminating the reliance on extensive offline fine-tuning or pre-collected datasets, our approach addresses the challenges of real-world deployment more directly.

\section{Method}
\label{sec:method}

\subsection{Optical flow from contrast maximization}

Each pixel in an event camera independently detects changes in brightness and generates an event $e_k = (t_k, \bm{x}_k, p_k)$ when this change exceeds a preset threshold. The polarity $p_k \in \{+1, -1\}$ indicates whether the brightness at pixel $\bm{x}_k$ and time $t_k$ increased or decreased. Contrast maximization~\cite{gallego2018unifying,gallego2019focus} assumes these events $\mathcal{E} = \{e_k\}_{k=1}^{N_e}$ are triggered by motion, meaning that a warp $e_k = (t_k, \bm{x}_k, p_k) \mapsto e'_k = (t_\text{ref}, \bm{x}'_k, p_k)$ with the correct motion estimate $\Delta\bm{x}_k$ will align it with other events triggered by the same portion of a moving edge, increasing the contrast of the image of warped events (IWE).

We follow the contrast maximization framework as in~\cite{paredes-valles2023taming}, which estimates optical flow for thin slices of the event stream using a recurrent architecture. By concatenating flows $\bm{u}_i$ as $\Delta\bm{x}_k=\sum_i(\Delta t_i \bm{u}_i)(e_k)$, events can be warped iteratively to neighboring slices, with the correct flows leading to sharp IWEs at all reference times along the trajectory:
\begin{equation}
    \mathcal{L}_\text{CM} = \frac{1}{T + 1} \sum_{t_\text{ref} = 0}^T \frac{\sum_k \bar{t}_k(t_\text{ref}) \kappa(\bm{x}_k)}{\sum_k \kappa(\bm{x}_k)}
    \label{eq:multiref}
\end{equation}

where $\bar{t}_k(t_\text{ref})$ is the timestamp contribution of event $e_k$ to the IWE at reference time $t_\text{ref}$, and $\kappa$ is a bilinear splatting kernel. We regularize (prevent event collapse) by scaling IWEs by the number of pixels with at least one event and by masking events that get warped out of the image space at any point~\cite{hagenaars2021selfsupervised,paredes-valles2023taming}.

\subsection{Combining depth and ego-motion into flow}

Assuming a static scene and no occlusion/disocclusion, depth and ego-motion can be accurately estimated from monocular video alone~\cite{zhou2017unsupervised}. The optical flow used to warp a pixel $\bm{x}$ between different views is constructed from depth $D$ and a camera transformation or relative pose $P$:
\begin{equation}
    \bm{x}' \sim KPD(\bm{x})K^{-1}\bm{x}
\end{equation}

with $K$ the camera intrinsic matrix, $P$ consisting of a rotation $R$ and a translation $\bm{t}$, and $\sim$ because depth is only defined up to a scale.\footnote{For simplicity, we omit the conversion to homogeneous coordinates.} The network estimates depth $D$ directly using a softplus activation~\cite{gordon2019depth}; relative pose $P$ is estimated with rotation expressed in exponential coordinates $\bm{\omega}$~\cite{wang2018learning}, and converted to $R$ using Rodrigues' formula.

To encourage consistent scale for consecutive depth predictions, we include the geometry consistency loss from~\cite{bian2019unsupervised}, which computes a normalized difference between the forward-projected depth $D_{0\shortrightarrow1}$ and interpolated depth $D'_1$ for all valid pixels $\bm{x}\in V$ (visible in both images):
\begin{equation}
    \mathcal{L}_\text{geo} = \frac{1}{\lvert V \rvert} \sum_{\bm{x}\in V} \frac{\lvert D_{0\shortrightarrow1}(\bm{x}) - D'_1(\bm{x}) \rvert}{D_{0\shortrightarrow1}(\bm{x}) + D'_1(\bm{x})}
\end{equation}

where we average over the number of valid pixels $\lvert V \rvert$. Setting an appropriate weight $\lambda$, this then results in the full loss formulation as:
\begin{equation}
    \mathcal{L} = \mathcal{L}_\text{CM} + \lambda\mathcal{L}_\text{geo}
\end{equation}

\begin{figure*}
    \centering
    \includegraphics[width=\linewidth]{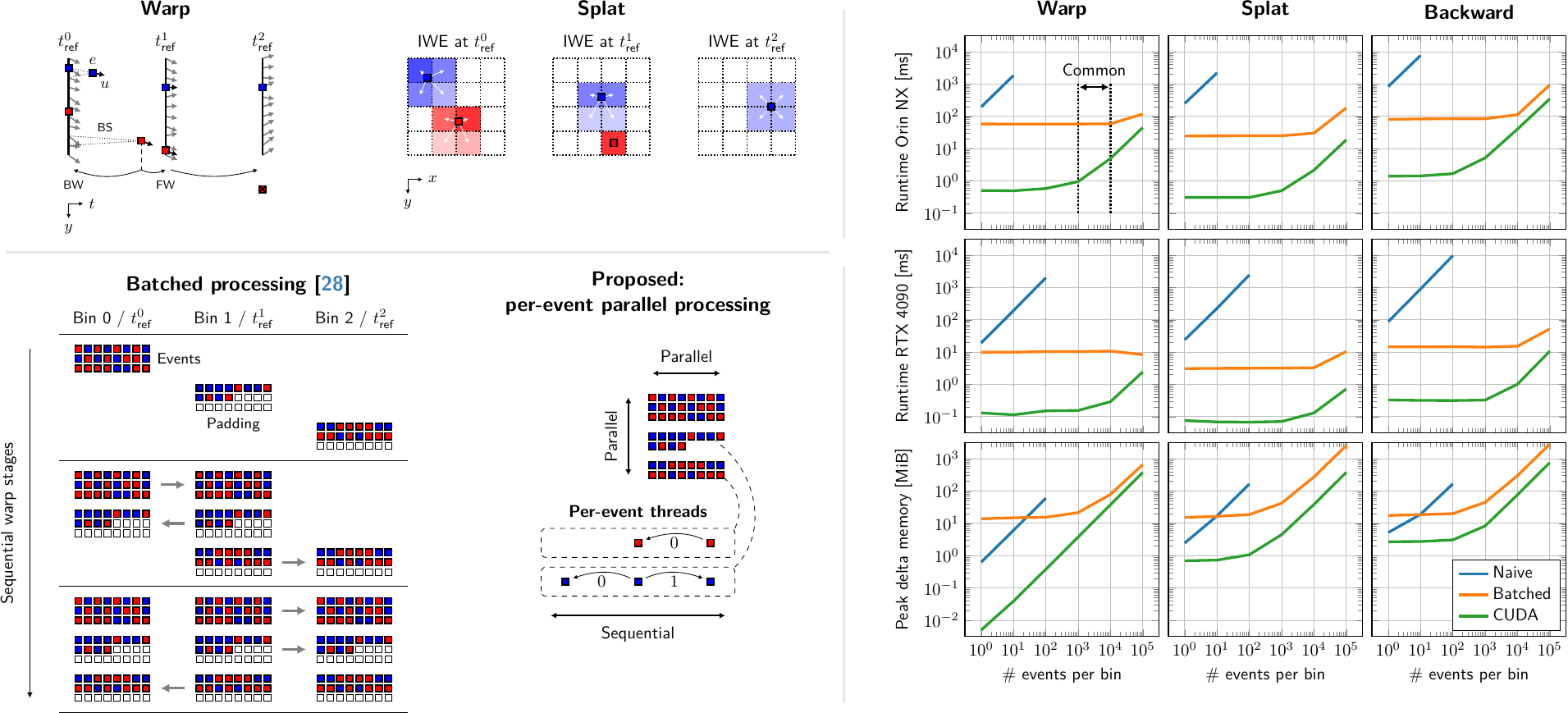}
    \caption{\textbf{Top left:} events $e$ of different polarities are warped forward and backward by bilinearly sampled (BS) optical flows $\bm{u}$. Events warped outside the image are discarded. Next, events are bilinearly splatted to IWEs (images of warped events) at all reference times $t^*_\text{ref}$. \textbf{Bottom left:} batched processing of events such as in~\cite{paredes-valles2023taming} requires zero-padding bins of events to equal length to facilitate simultaneous warping to neighboring reference times. In contrast, our per-event parallel processing in CUDA warps all events independently, doing away with padding and allowing to warp only those events still in the image space. \textbf{Right:} Runtime and peak increase in memory consumption for different phases of computing the contrast maximization loss on an NVIDIA RTX 4090 and Jetson Orin NX. 10\% of each bin is made up of padding, which is not processed by the CUDA implementation. We indicate the range of events per bin for common datasets~\cite{delmerico2019are,zhu2018multivehiclea} in black. Naive PyTorch processes all events in a for-loop. While batching events together improves a lot over this, parallel processing of all events in CUDA results in even bigger speedups with less memory consumed.}
    \label{fig:cuda}
\end{figure*}

\subsection{Optimizations for on-device learning}

For efficient prediction, we make use of a lightweight, 430k-parameter network inspired by~\cite{wu2024lightweight}. It consists of a strided convolutional encoder, a ConvGRU recurrent bottleneck, and a two-tailed convolutional decoder for depth and ego-motion (more details in the supplementary material). While one network forward pass takes less than a millisecond on an NVIDIA RTX 4090, computing the contrast maximization loss and backpropagating the resulting gradients each take up more than 10~ms.

To make on-device learning feasible, we have to improve the efficiency of the components that make up the loss computation and network update: i) warping all events in the accumulated set of events $\mathcal{E}$ using a sampled optical flow to all reference times $t_\text{ref}\in [0, T]$, ii) bilinearly splatting them to the IWE at that $t_\text{ref}$, iii) computing the gradient with respect to the network parameters.

Previous work~\cite{paredes-valles2023taming} warped and splatted events in batches using PyTorch functions. This has multiple inefficiencies. Batching different amounts of events together leads to padding with zeros, resulting in wasteful computation and memory usage. This also goes for warping events that already went out of the image space (and therefore do not contribute to the loss anymore). Furthermore, some operations (like bilinear splatting) do not have optimized implementations in PyTorch. All this combined results in extra computational and memory overhead due to intermediate tensor allocations, multiple instead of single kernels, computation graph tracking and scattered memory accesses.

The abovementioned issues can be resolved by considering that all events independently contribute to the loss since they are summed in the IWEs, and that we can therefore parallelize over all the accumulated events $\mathcal{E}$. We implement the functions to do so in CUDA, getting rid of most of the overhead, and connect them to PyTorch as an extension. 

Specifically, we assign one CUDA thread to handle one or more events in parallel, read off their positions in the 3D domain $(x,y,t)$, and then apply the flow fields to ``push'' each event through time either forward or backward in an iterative fashion. During the backward pass, we use the stored warped-point positions along each time step (forward and backward) to compute partial derivatives w.r.t the flow vector $\frac{\partial \mathcal{L}}{\partial \mathbf{f}}$. As in the forward pass, we rely on bilinear interpolation weights at each $(x,y)$ to distribute gradients to the four nearest flow-vector cells. By avoiding padding and unnecessary warping computations, we achieve significantly better parallelization on GPUs.

As shown in \cref{fig:cuda}, the resulting improvements are, depending on the device, roughly 100x in terms of runtime, and 2-5x in terms of memory consumption. Common datasets like UZH-FPV~\cite{delmerico2019are} and MVSEC~\cite{zhu2018multivehiclea} have between 1k and 10k events per bin, well within the range of these speedups. Furthermore, looking at the peak delta memory, the removal of padding yields much lower memory consumption when event data is highly sparse. Because we now provide the analytical gradients in the CUDA backward kernel, these do not have to be computed through automatic differentiation, leading to further efficiency improvements.

\subsection{Using depth for obstacle avoidance}

In the absence of metric depth and with a possibly varying scale, we can construct simple obstacle-avoiding behaviour using the difference in predicted depths for different parts of the field-of-view~\cite{liu2023nano,chakravarty2017cnnbased}. More specifically, we slice the depth map into $K$ vertical bins, compute the average inverse depth $d_k$ for each, and use these to set a desired yaw rate $\dot{\psi}$:
\begin{align}
    \dot{\psi} &= \dot{\psi}_\text{goal}(\bm{d}) + \dot{\psi}_\text{avoid}(\bm{d}) \\
    \dot{\psi}_\text{goal}(\bm{d}) &= \lambda_\text{goal} (\argmin_k (d_k) - \bar{k}) \\
    \dot{\psi}_\text{avoid}(\bm{d}) &= \lambda_\text{avoid} \sum_{k=0}^{K-1} (\bar{k} - k) e^{(-\frac{\alpha}{d_k})}e^{(-\frac{(k - \bar{k})^2}{2\sigma^2})}
\end{align}

where yawing to the right is positive, and $\bar{k}=\frac{K-1}{2}$ is the center index. The resulting behaviour is both obstacle-avoiding ($\text{avoid}$ part) and depth-seeking ($\text{goal}$ part).

\begin{figure}
    \centering
    \includegraphics[width=\linewidth]{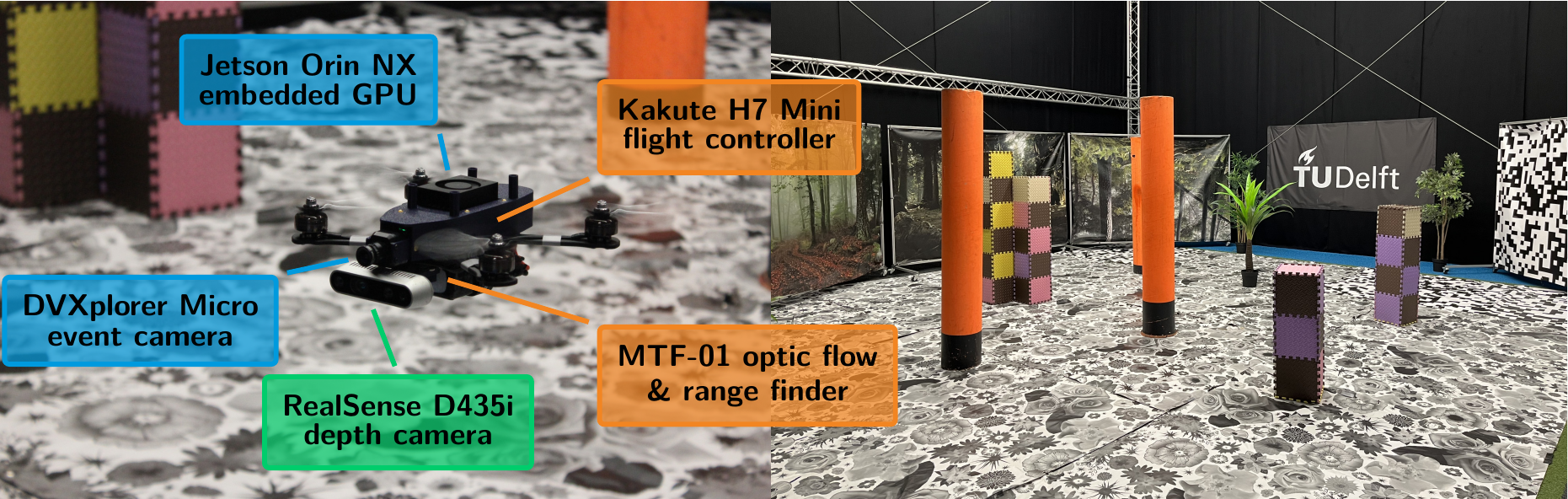}
    \caption{Overview of the drone (left) and the flight environment (right). System components in blue are for the on-board depth learning pipeline, orange components are for low-level flight control, and green components are for logging only.}
    \label{fig:drone}
\end{figure}

\subsection{Drone and flight environment}

The experimental setup as shown in \cref{fig:drone} consists of a custom 5-inch quadrotor with a total weight of approximately 800~g, including all sensors, actuators, on-board compute and battery. All algorithms are implemented to run entirely on board, using an NVIDIA Jetson Orin NX embedded GPU to receive data from the event camera, perform learning and estimate depth in real time. Control commands (yaw rate) based on the predicted depth maps are sent to the flight controller, a Kakute H7 Mini running the open-source autopilot software PX4. Communication between PX4 and the Orin is done using ROS2~\cite{macenski2022robot}. An MTF-01 optic flow sensor and rangefinder enables stable autonomous flight where only yaw rate is controlled based on the depth estimate.

To keep the event rate down (below 1~Mev/s), we only turn on every fourth pixel on the DVXplorer Micro, resulting in a 160x120 stream (instead of 640x480) for the same field-of-view. These events are accumulated into 20~ms windows and made into a frame for the network. The whole events-to-depth pipeline is running at approximately 30~Hz while learning, consuming on average around 9~W. We include a RealSense D435i depth camera for logging purposes only. To plot ground-truth flight trajectories, we record the drone's position using a motion capture system.

\section{Experiments}
\label{sec:experiments}

\subsection{Event-based depth benchmarks}

\begin{figure*}[!ht]
    \centering
    \includegraphics[width=\linewidth]{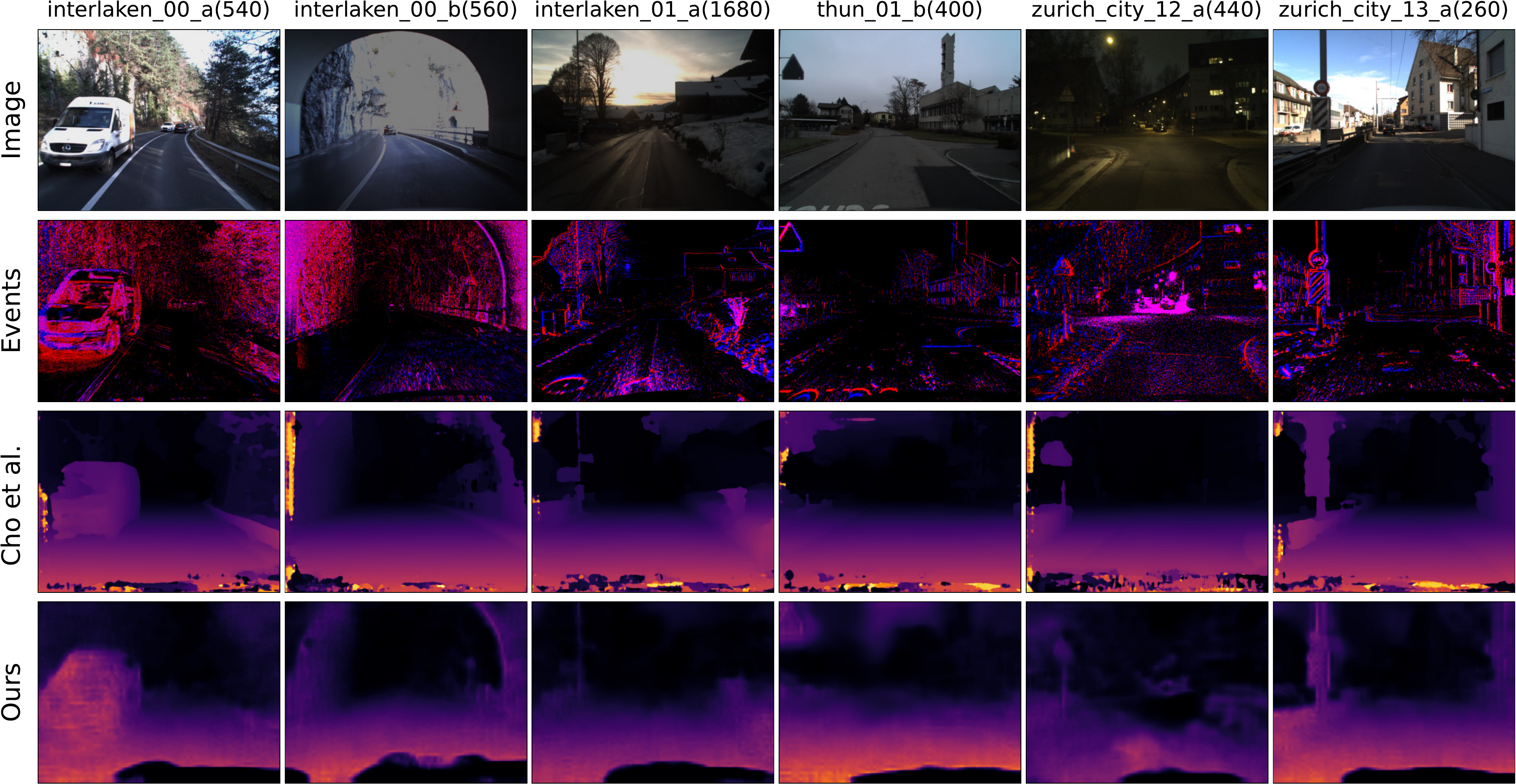}
    \caption{Qualitative results of disparity predictions on the DSEC disparity benchmark. Images are for visualization only, as disparity estimation is event-based. The same color map is applied to the disparity values from the stereo- and supervised-learning-based method from Cho~\textit{et al.}~\cite{cho2025temporal} and our monocular, self-supervised learning method for easy comparison.}
    \label{fig:dsec}
\end{figure*}

\paragraph{Setup.} We train our proposed network on the training sets with a batch size of 8 and a constant learning rate of 1e-4 with the Adam optimizer for 50 epochs (more details in the supplementary material). We do truncated backpropagation through time, with a backward pass/gradient update conducted every 10 forward passes. Detaching the network while not resetting its state ensures bounded memory usage, mitigates potential gradient explosion/vanishing, and allows the network to retain temporal context effectively. The quantitative evaluations on MVSEC and DSEC are provided in \cref{table:mvsec-mae} and \cref{table:dsec}.

\begin{table}
\small

\begin{center}

\begin{tabular}{lcccccc}
\toprule
     &  \multicolumn{3}{c}{\texttt{outdoor\_day1}} & \multicolumn{3}{c}{\texttt{outdoor\_night1}}\\
      \cmidrule(lr){2-4} \cmidrule(lr){5-7} 
     Depth cutoff & 10m & 20m & 30m & 10m & 20m & 30m \\
     \midrule
     \rowcolor{lightgray} Zhu~\textit{et al.}~\cite{zhu2023selfsupervised}\,\tablefootnote{The network uses events as input but was trained with intensity frames in the loss function. Therefore, it is included as a reference but is not directly comparable to self-supervised methods trained solely on event data.} & {1.40} & {2.07} & {2.65} & {2.18} & {2.70} & {3.64}\\
     Zhu~\textit{et al.}~\cite{zhu2023selfsupervised} & 3.90 & \underline{3.79} & 4.89 & 5.55 & 4.57 & 5.72\\
     Zhu~\textit{et al.}~\cite{zhu2019unsupervised} & \underline{2.72} & 3.84 & \underline{4.40} & \textbf{3.13} & \underline{4.02} & \underline{4.89}\\
     \textbf{Ours} &  \textbf{2.25} & \textbf{3.36} & \textbf{4.23} & \underline{3.25} & \textbf{3.83} & \textbf{4.50} \\
     \hdashline
     \textbf{Ours (dense)} & 1.96 & 2.67 & 3.29 & 2.92 & 3.56 & 4.28\\
\bottomrule
\end{tabular}
\vspace{-0.5cm}
\end{center}
\caption{MAE (mean absolute error) of depth prediction in meters on MVSEC test sequences at various depth cutoff distances. The best result is highlighted in bold, and the second best is underlined. The method shown in the shaded row serves as a reference and is not directly comparable to the others, as it also uses image frames.}
\label{table:mvsec-mae}
\end{table}

\begin{table}
\small
\setlength{\tabcolsep}{4.5pt}
\begin{center}

\begin{tabular}{llcccc}
\toprule
      & & 1PE & 2PE & MAE & RMSE\\
     \midrule
     \multirow{2}{*}{\centering SL}
     & Cho~\textit{et al.}~\cite{cho2025temporal} &  8.966 & 2.345 & 0.501 & 1.175 \\
     & DSEC baseline~\cite{gehrig2021dsec} & 10.92 & 2.905 & 0.576 & 1.381 \\
     \hdashline
     \multirow{2}{*}{\centering SSL}
     & \textbf{Ours (best scale)} & 82.64 & 66.57 & 4.583 & 5.937\\
     & \textbf{Ours (approx. scale)} & 84.92 & 70.47 & 4.946 & 6.274\\
\bottomrule
\end{tabular}
\vspace{-0.5cm}
\end{center}
\caption{Quantitative evaluation on the DSEC disparity benchmark. Due to the lack of other monocular SSL (self-supervised learning) methods on the leaderboard, we compare against two representative stereo-based SL (supervised learning) methods.}
\label{table:dsec}
\end{table}

\paragraph{Results.} On MVSEC, our method outperforms the other two self-supervised, events-only baselines \cite{zhu2019unsupervised, zhu2023selfsupervised}. For context, we also provide the results from the approach in \cite{zhu2023selfsupervised}, which additionally uses intensity frames in the training process for a photometric consistency loss. Although it achieves a higher accuracy, our networks rely solely on event streams during training. For completeness, we also assess the accuracy of dense depth (i.e., not masked by events), as shown in the last row of \cref{table:mvsec-mae}.

In the absence of self-supervised methods on the DSEC disparity benchmark, we compare our approach against two top-performing stereo-event-based supervised learning baselines \cite{cho2022selection,gehrig2021dsec}. To convert our monocular unnormalized depth predictions from our network output into metric depth, we apply a scaling factor derived from the ratio of the median predicted depth to the ground truth median from the training set, labeled as ``approx. scale'' in \cref{table:dsec}. Additionally, we conduct a grid search on the scaling factor to achieve the highest accuracy on the test set, reported as ``best scale''.

While our accuracy on the DSEC disparity benchmark falls short of supervised baselines, qualitative comparisons in \cref{fig:dsec} demonstrate that our approach effectively captures meaningful structures within disparity maps, even without ground truth labels during training. Notably, close objects, such as the car in \texttt{interlaken\_00\_a(540)} and traffic signs in \texttt{thun\_01\_b(400)}, \texttt{interlaken\_01\_a(1680)} and \texttt{zurich\_city\_12\_a(400)}, are accurately represented. Although the boundaries in our results may lack the sharpness achieved by supervised baselines, our approach better preserves contour shapes, such as the front of the car in \texttt{interlaken\_00\_a(540)}, the arc of the tunnel in \texttt{interlaken\_00\_b(560)} and the pole in \texttt{zurich\_city\_13\_a(260)}. This advantage is especially evident for thin objects like the sign pole in \texttt{interlaken\_01\_a(1680)} and \texttt{zurich\_city\_12\_a(440)}, which are often challenging for supervised methods to capture accurately. Additionally, our self-supervised approach can run at higher-than-ground-truth frequencies (100~Hz vs 10~Hz) and is immune to artifacts typically caused by the sporadic availability of ground truth at the image boundaries, resulting in smoother disparity maps free from discontinuity artifacts.

\paragraph{Limitations.} Several factors constrain the accuracy of our methods on these benchmarks, including the reliance on self-supervised learning with events only, a compact network architecture, the use of monocular depth estimation rather than stereo and the imperfect estimation of a scaling factor for converting monocular depth to metric depth. Errors in few-event areas could be reduced by, e.g., including the reconstruction loss from~\cite{paredes-valles2021back}. However, our aim is not to surpass state-of-the-art methods, and we believe the quality of our depth predictions is sufficient to support downstream tasks like robot navigation.

\subsection{Drone experiments}

\paragraph{Setup.} We first pre-train our network on the UZH-FPV dataset~\cite{delmerico2019are} using our self-supervised pipeline. This dataset was chosen for its diverse set of motion trajectories, enabling the network to learn a latent representation that generalizes well across various motion types. After pre-training, the network is deployed on a drone, where online learning (fine-tuning) is performed during flight. The network's forward pass operates at an average speed of 30~Hz, with a backward pass and gradient update conducted every 10 forward passes. During flight experiments, we set the drone to fly at a constant height and a forward speed of 0.5~m/s. The predicted depth is binned and used to control the drone's yaw rate.

\begin{figure}[t]
    \centering
    \includegraphics[width=\linewidth]{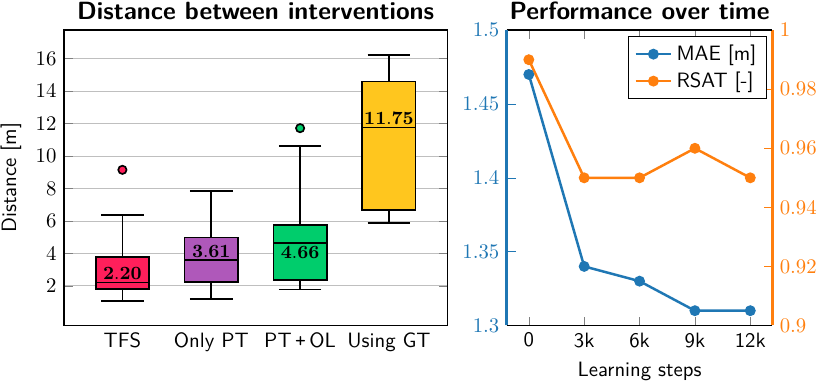}
    \caption{\textbf{Left:} Boxplots of distance between pilot interventions during flight experiments. While using ground truth (GT) depth is best, adding online learning (PT + OL) improves over just pre-training (PT) by \textasciitilde{}30\%. Training from scratch (TFS) does not result in meaningful obstacle avoidance. \textbf{Right:} MAE (mean absolute error) of depth prediction and RSAT (ratio of squared average timestamps, indicates deblurring quality) during online learning in flight. Model checkpoints were saved periodically and evaluated on a test sequence unseen by the model beforehand. 300 learning steps correspond to roughly 100 seconds of training during flight.}
    \label{fig:finetune-metrics-curve}
    \vspace{-0.3cm}
\end{figure}

\paragraph{Results.} We show the quantitative improvements achieved through online learning in \cref{fig:finetune-metrics-curve} (right plot), where saved checkpoints are evaluated on a test sequence recorded in the same environment but with different placements for obstacles. The model shows significant improvement not only in the RSAT (ratio of squared average timestamps) metric~\cite{hagenaars2021selfsupervised}, which is strongly correlated with the contrast maximization loss used to optimize the network, but also in MAE (mean absolute error) when compared against ground truth depth. Additionally, the fine-tuning process is efficient, converging within just two minutes of flight.

\begin{figure}[t]
    \centering
    \includegraphics[width=\linewidth]{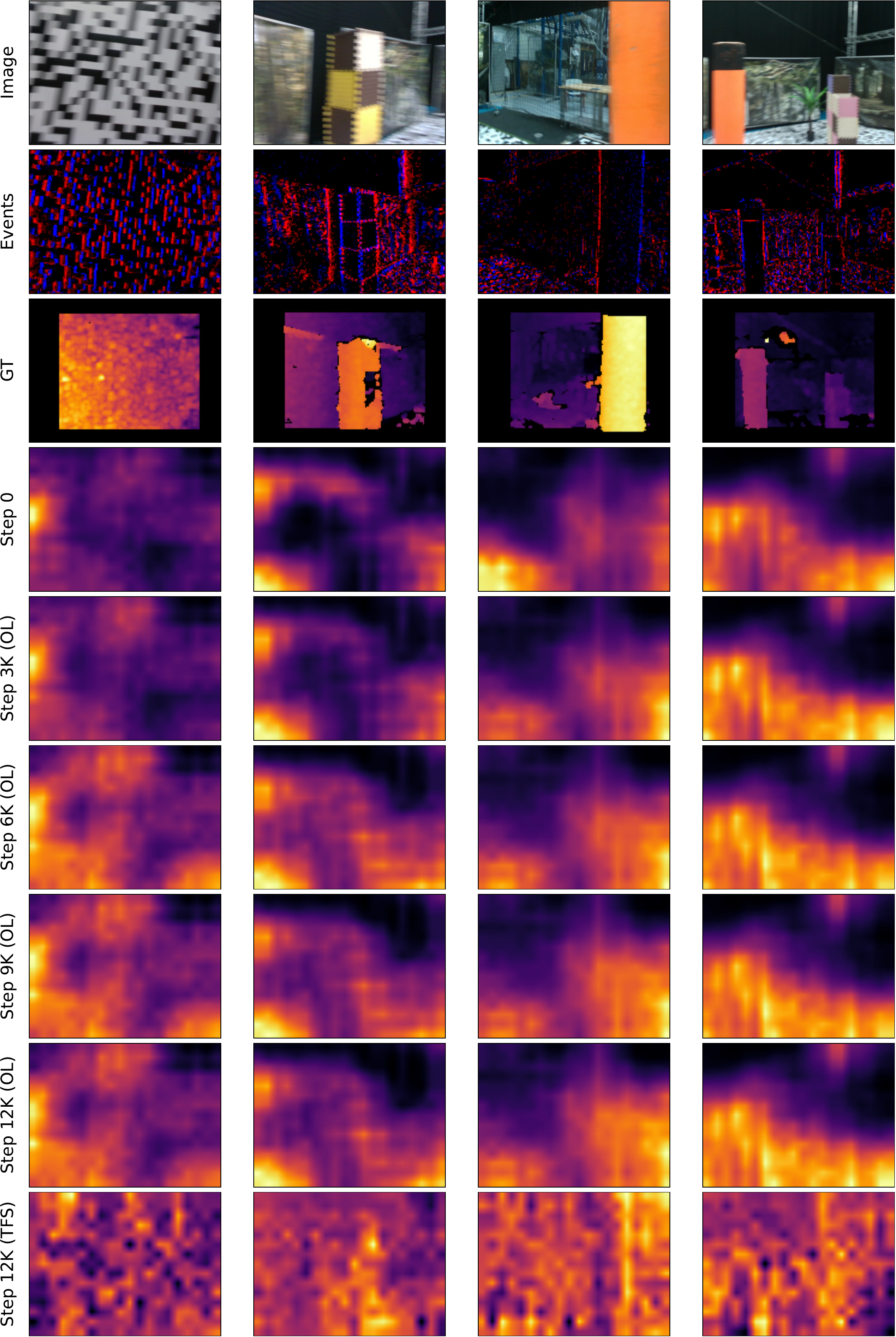}
    \caption{Qualitative visualization of disparity map evolution during online learning. Note that the image is for visualization purposes only, as disparity estimation is event-based. The same color map is applied to predictions from all different models for easy comparison. The pre-trained network begins at step 0, followed by 12K steps of online learning (OL) with streaming event data during flight. For comparison, we also show the prediction quality of a randomly initialized network trained from scratch (TFS) for 12K steps.}
    \label{fig:finetune-qual}
\end{figure}

Qualitative results at different snapshots during online learning are presented in \cref{fig:finetune-qual}. Compared to step 0 (the pre-trained model), the disparity values for certain close objects, such as the wall and poles, increase, as evidenced by the brightened colors in those regions. To highlight the benefits of pre-training, we also compare our model with a network initialized with random weights and trained using the same amount of online learning data, as shown in the last row of \cref{fig:finetune-qual}. The from-scratch network fails to produce meaningful disparity maps within the flight's limited timespan, underscoring the importance of pre-training in achieving fast and reliable adaptation during online learning.

Finally, we quantify that these improvements in depth estimation through online learning translate to better obstacle avoidance performance in flight experiments. During each experiment, the drone takes off and immediately starts to fly autonomously. A human pilot monitors the flight and intervenes if the drone is in a near collision with an obstacle. After the human pilot corrects for the collision course, the drone is switched to autonomous flying again. 

\begin{figure*}
    \centering
    \includegraphics[width=\linewidth]{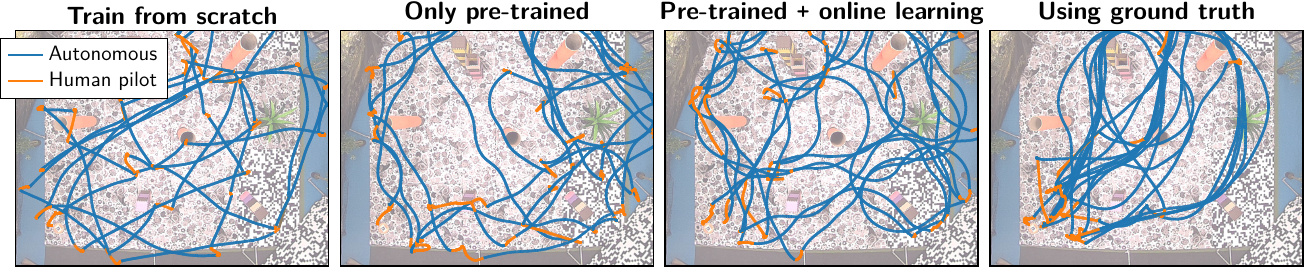}
    \caption{Top-view flight trajectories of various experiments. Blue represents autonomous flight, while orange indicates pilot interventions necessary to prevent collisions/going out of bounds. Training from scratch does not give meaningful obstacle-avoiding behavior. Online learning results in longer autonomous sections and more diverse paths than just pre-training. Using ground truth depth (from RealSense) results in almost-perfect avoidance, and only requires intervention when flying into a corner (limitation of control algorithm).}
    \label{fig:trajectory}
\end{figure*}

We compare the distance between pilot interventions for the different experiments in \cref{fig:finetune-metrics-curve} (left plot), and show top views of the flight trajectories in \cref{fig:trajectory}. When training from scratch (network initialized with random weights), the drone does not avoid obstacles and mostly flies in straight lines. When we start with a pre-trained network, we see actual obstacle-avoiding behavior, and the distance between pilot interventions goes up by \textasciitilde{}65\%. When adding online learning during flight, the distance between interventions improves by a further \textasciitilde{}30\%, and the flight behavior seems to become more diverse. 

\paragraph{Limitations.} The quality of the on-board depth maps is limited by the fact that only a quarter of the camera's 640x480 resolution is used (compare \cref{fig:dsec} and \cref{fig:finetune-qual}). We mitigate this with an artificially textured environment to ensure sufficient motion-induced events. Higher-quality depth maps, or operation in more natural environments, will require using more of the camera's resolution. 

To further enhance computational efficiency and performance, the nonlinear motion model~\cite{paredes-valles2023taming} could be traded for the cheaper-to-compute linear variant~\cite{hagenaars2021selfsupervised}, at the cost of increased errors on nonlinear event trajectories. Furthermore, inference and learning could be run asynchronously at different rates~\cite{vodisch2023covio}, or only limited to partial network fine-tuning for a few selected layers. Incorporating a jointly optimized flow decoder tail~\cite{yin2018geonet} could also improve depth estimation for dynamic obstacles.

Dynamic objects moving towards the drone would already be avoided by our current pipeline (even though they are not included in training). However, due to the static scene assumption, their depth is underestimated (like the oncoming van in \cref{fig:dsec}).

Lastly, The current depth-based yaw control is attracted by corners in the environment, requiring pilot intervention (see the bottom left environment corners in \cref{fig:trajectory}). Solving this, or allowing for more complex environments (e.g., higher obstacle density), would require a more advanced control strategy capable of better interpreting depth cues.

\section{Conclusion}
\label{sec:conclusion}

We have improved the efficiency of self-supervised learning of monocular depth estimation from events, such that on-device learning of low-latency monocular depth and ego-motion becomes feasible. The proposed approach features more efficient and parallel processing, and has been implemented in CUDA instead of PyTorch. For common event rates (0.1-1~Mev/s) this reduces runtime by 100x, while using 2-5x less memory---improvements that would also transfer to other pipelines involving warping/splatting of events~\cite{shiba2024secrets,falanga2020dynamic,paredes-valles2021back} and images~\cite{niklaus2023splattingbased}. 

When trained and benchmarked on event camera datasets, our small recurrent network  outperforms other self-supervised approaches and captures essential structures with sufficient quality to support downstream navigation tasks. Furthermore, we demonstrate that online learning on board a small flying drone leads to improved depth estimates within two minutes of learning, leading to more successful obstacle avoidance (\textasciitilde{}30\% improvement in distance between pilot interventions).

Our work taps into the unused potential of on-board, online self-supervised learning. The current results already demonstrate that online learning leads to better performance in the operational environment.  While SSL still needs further improvements to match supervised baselines, its core advantages---pretraining on large unlabeled datasets and finetuning directly in the test environment---hold the key to truly robust autonomous robot deployment across diverse real-world settings.

\paragraph{Acknowledgments.} The authors would like to thank the reviewers for their constructive feedback and suggestions. This work was supported by funding from NWO (NWA.1292.19.298), the Air Force Office of Scientific Research (award no. FA8655-20-1-7044) and the Office of Naval Research Global (award no. N629092112014).

\newpage
{
    \small
    \bibliographystyle{ieeenat_fullname}
    \bibliography{main}
}

\clearpage
\setcounter{page}{1}
\def\maketitlesupplementary
   {
   \newpage
       \twocolumn[
        \centering
        \Large
        \textbf{\thetitle}\\
        \vspace{0.5em}Supplementary Material \\
        \vspace{0.2em}\large\url{https://mavlab.tudelft.nl/depth_from_events}
        \vspace{1em}
       ] %
   }
\maketitlesupplementary

\section{Extra qualitative results}

\begin{figure*}
    \centering
    \includegraphics[width=\linewidth]{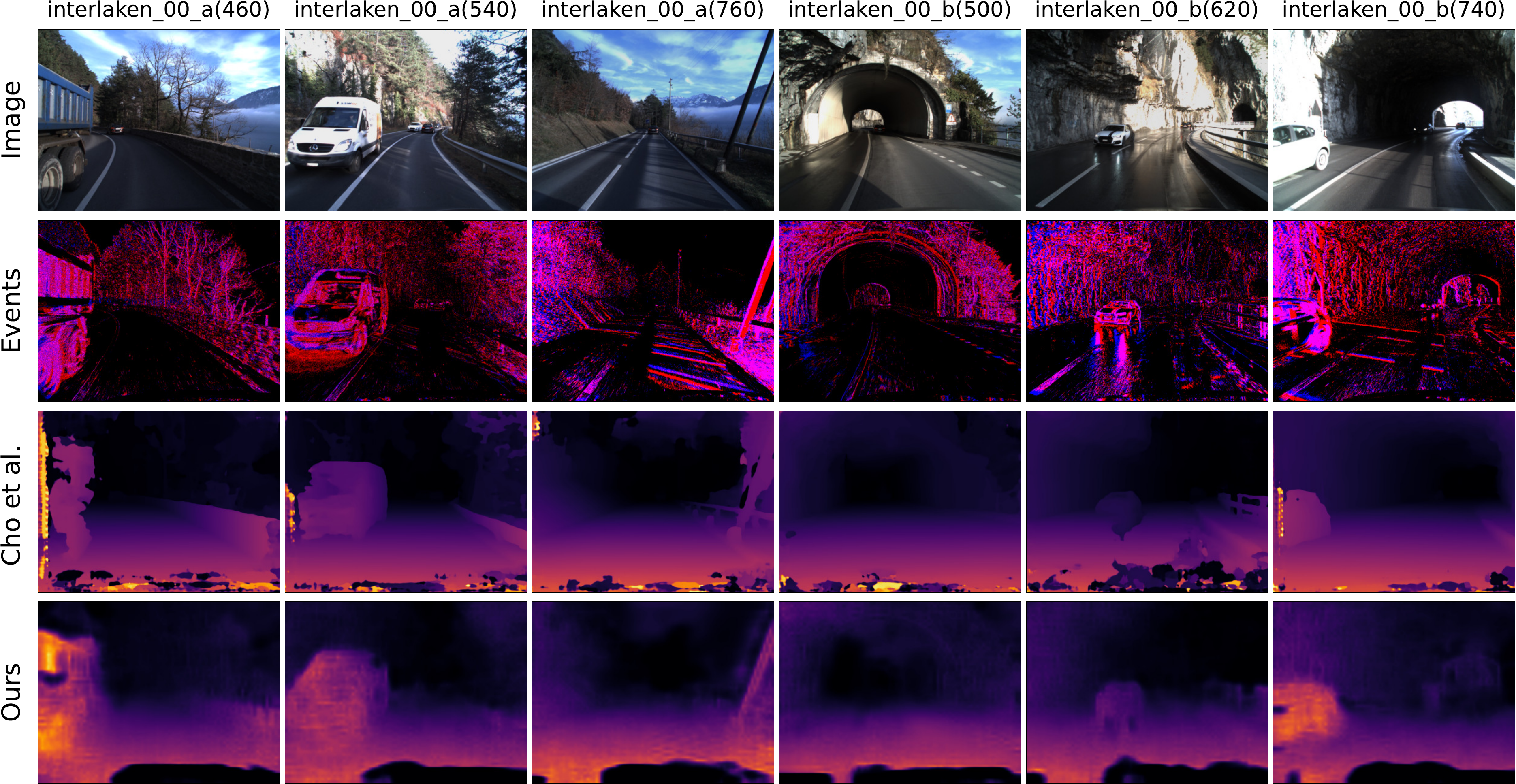}\\[1em]
    \includegraphics[width=\linewidth]{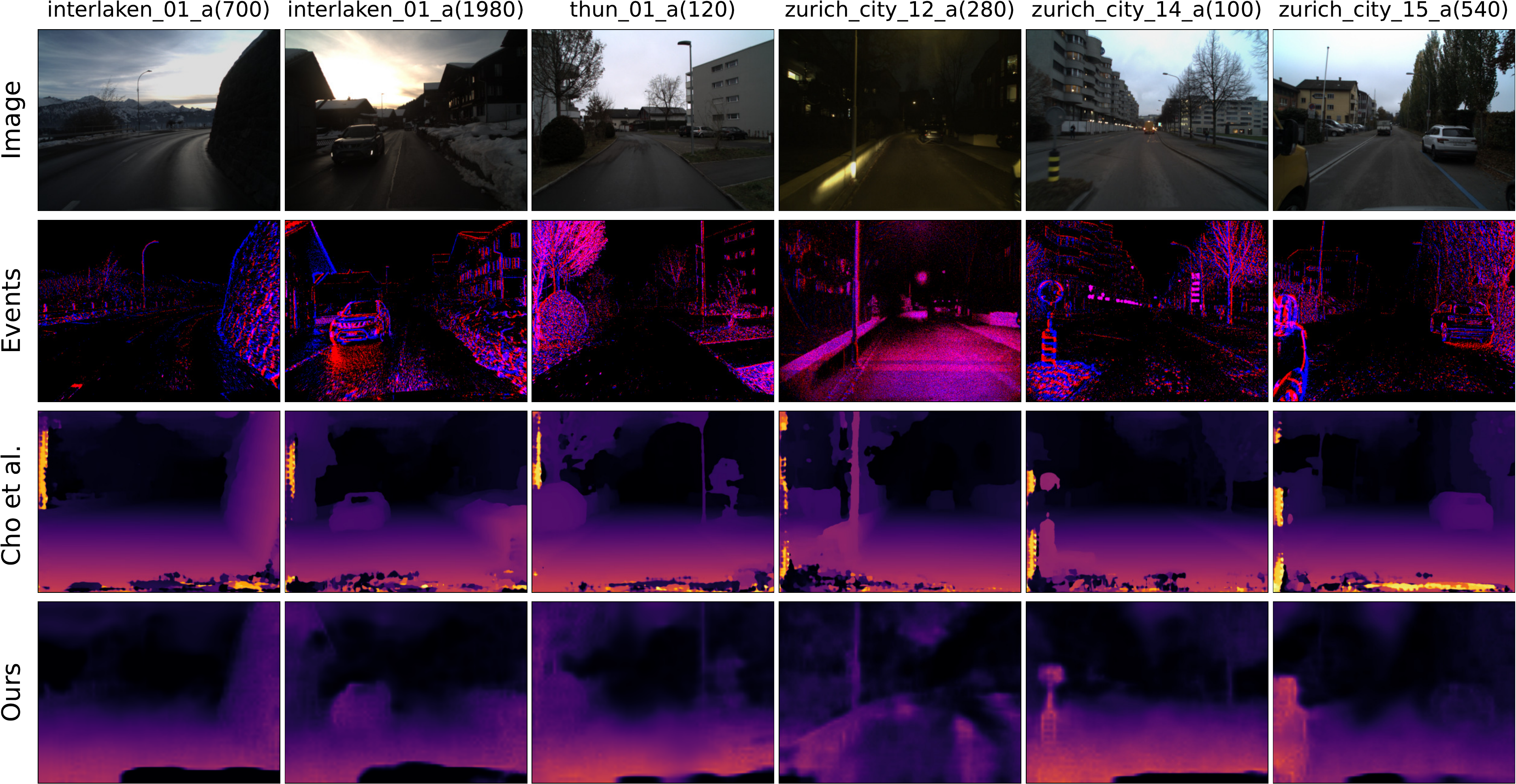}
    \caption{Additional qualitative results of disparity predictions on the DSEC disparity benchmark. Images are for visualization only, as disparity estimation is event-based. The same color map is applied to the disparity values from the stereo- and supervised-learning-based method from Cho~\textit{et al.}~\cite{cho2025temporal} and ours for easy comparison.}
    \label{fig:extra-dsec-qual}
\end{figure*}

\paragraph{DSEC.} We present additional qualitative results from the DSEC test set in \cref{fig:extra-dsec-qual}. While the self-supervised results exhibit less sharp boundaries, they are free from artifacts commonly introduced by supervised learning, such as discontinuities at image borders and around thin objects.

\begin{figure*}
    \centering
    \includegraphics[width=\linewidth]{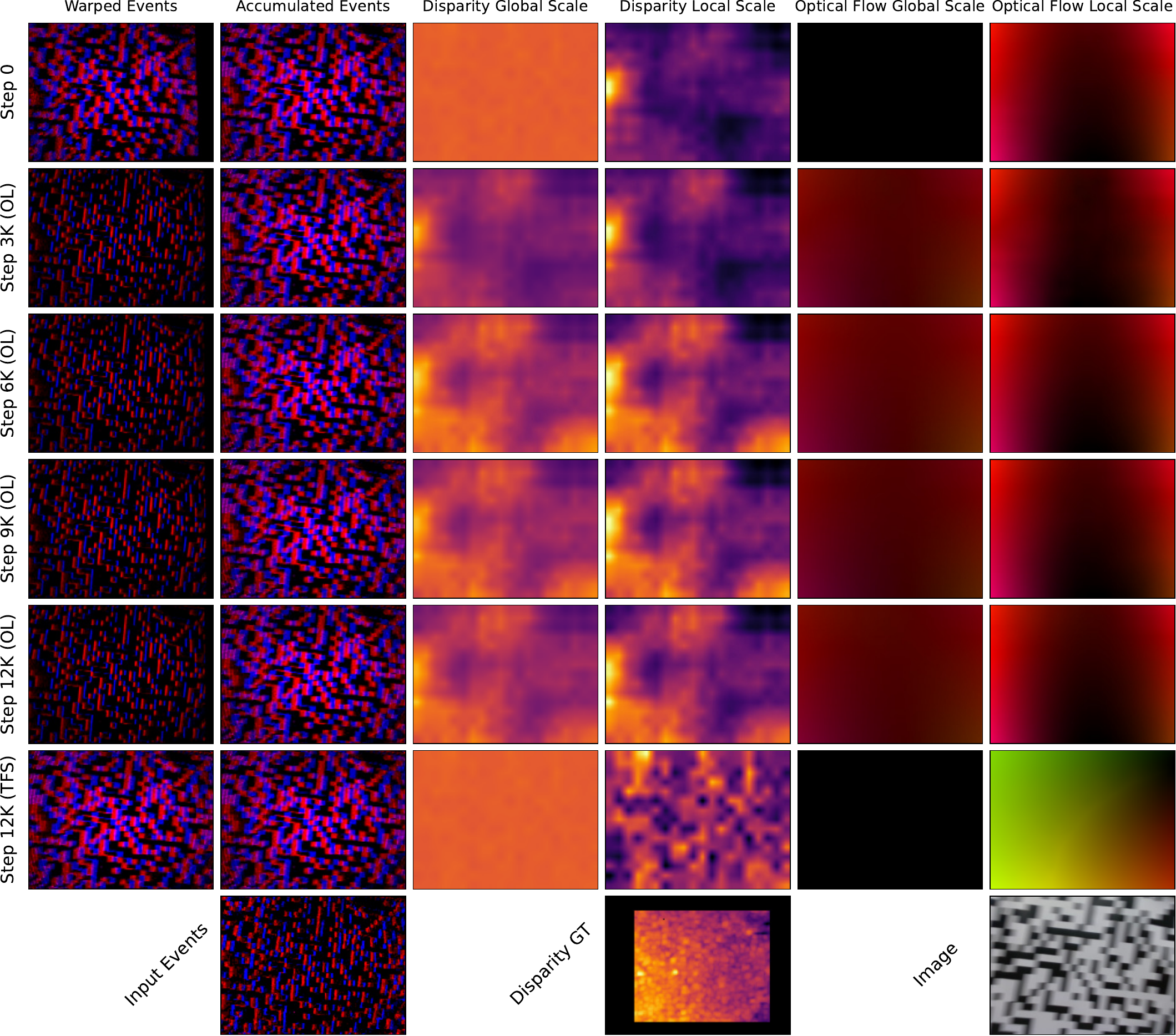}
    \caption{Extended qualitative results on unseen data from a flight test recording. From top to bottom, we evaluate a pre-trained-only network (``step 0''), then four networks after increasing amounts of online learning (OL), and finally a network trained-from-scratch. The bottom row starts with the single 20~ms bin of events currently seen by the network. The rest of the second column shows the accumulated events (multiple bins) in the current contrast maximization loss window. Applying the iterative warp by the optical flow constructed from depth and ego-motion gives the warped and deblurred events in the first column. Columns three and four show disparity at a global (color map shared between rows) and local (color map for only that row) scale. The last two columns show optical flow constructed from depth and ego-motion with global and local color maps, respectively.}
    \label{fig:suppqual2473}
\end{figure*}

\begin{figure*}
    \centering
    \includegraphics[width=\linewidth]{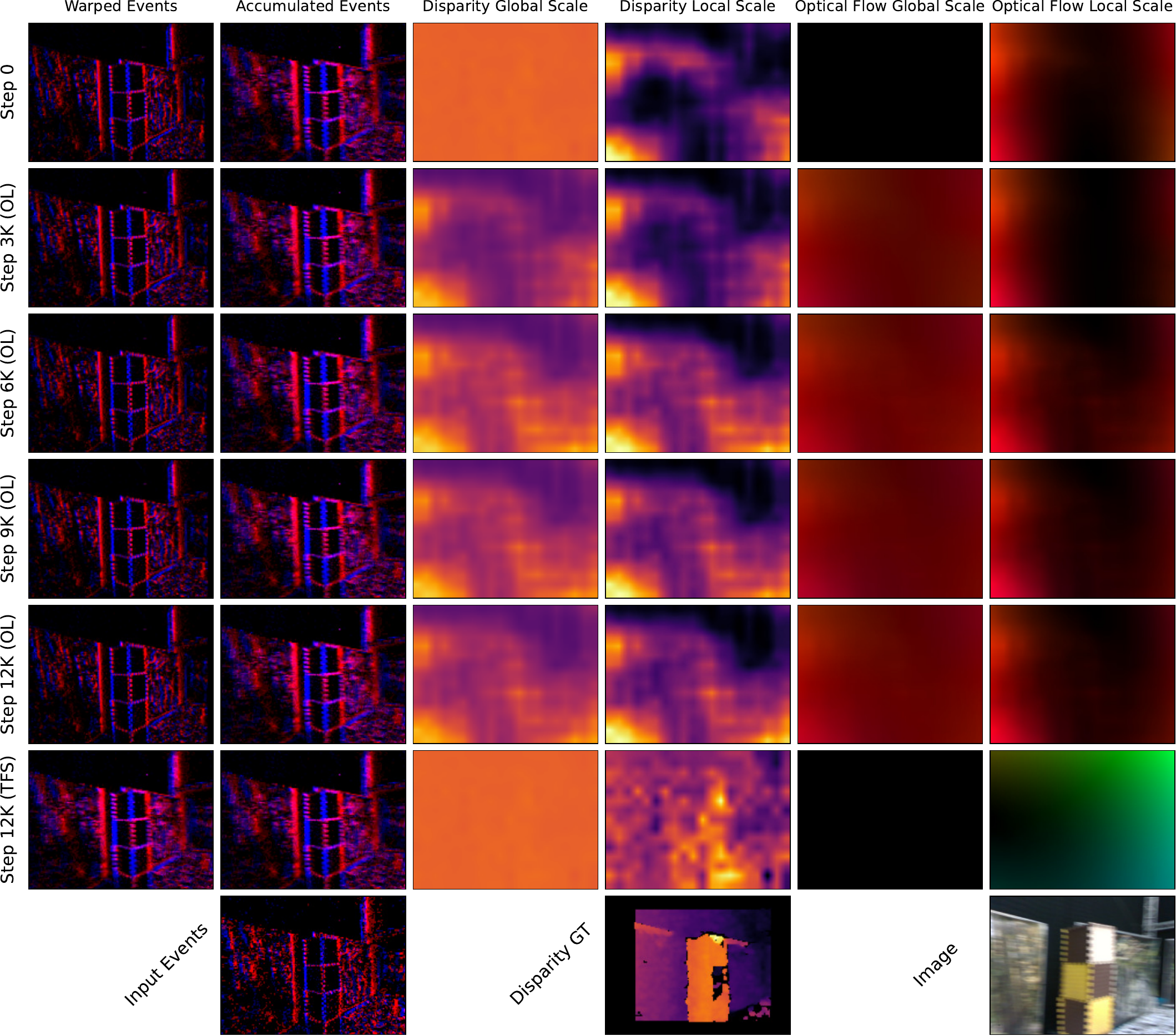}
    \caption{Extended qualitative results on unseen data from a flight test recording. From top to bottom, we evaluate a pre-trained-only network (``step 0''), then four networks after increasing amounts of online learning (OL), and finally a network trained-from-scratch. The bottom row starts with the single 20~ms bin of events currently seen by the network. The rest of the second column shows the accumulated events (multiple bins) in the current contrast maximization loss window. Applying the iterative warp by the optical flow constructed from depth and ego-motion gives the warped and deblurred events in the first column. Columns three and four show disparity at a global (color map shared between rows) and local (color map for only that row) scale. The last two columns show optical flow constructed from depth and ego-motion with global and local color maps, respectively.}
    \label{fig:suppqual3233}
\end{figure*}

\begin{figure*}
    \centering
    \includegraphics[width=\linewidth]{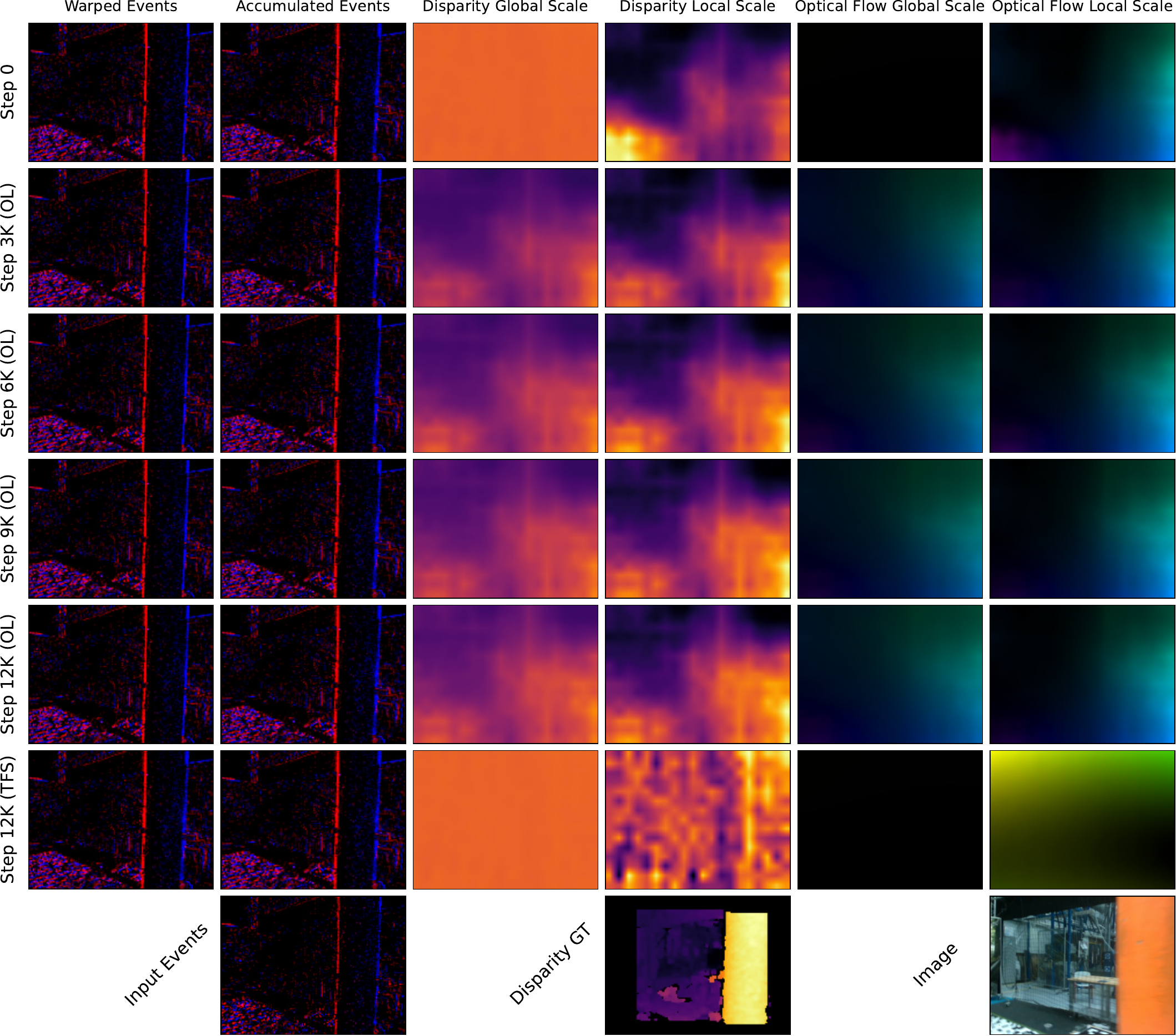}
    \caption{Extended qualitative results on unseen data from a flight test recording. From top to bottom, we evaluate a pre-trained-only network (``step 0''), then four networks after increasing amounts of online learning (OL), and finally a network trained-from-scratch. The bottom row starts with the single 20~ms bin of events currently seen by the network. The rest of the second column shows the accumulated events (multiple bins) in the current contrast maximization loss window. Applying the iterative warp by the optical flow constructed from depth and ego-motion gives the warped and deblurred events in the first column. Columns three and four show disparity at a global (color map shared between rows) and local (color map for only that row) scale. The last two columns show optical flow constructed from depth and ego-motion with global and local color maps, respectively.}
    \label{fig:suppqual4972}
\end{figure*}

\begin{figure*}
    \centering
    \includegraphics[width=\linewidth]{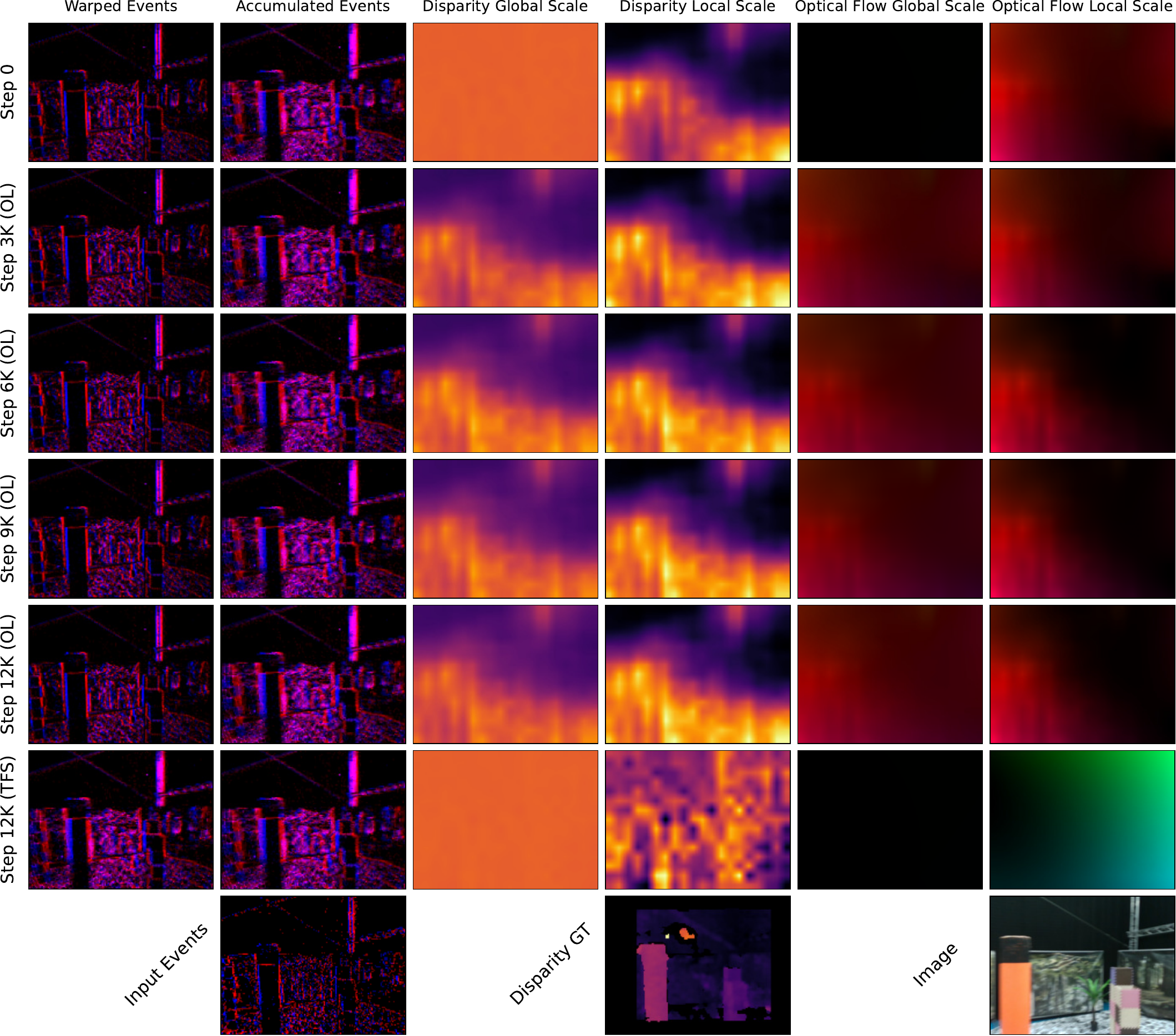}
    \caption{Extended qualitative results on unseen data from a flight test recording. From top to bottom, we evaluate a pre-trained-only network (``step 0''), then four networks after increasing amounts of online learning (OL), and finally a network trained-from-scratch. The bottom row starts with the single 20~ms bin of events currently seen by the network. The rest of the second column shows the accumulated events (multiple bins) in the current contrast maximization loss window. Applying the iterative warp by the optical flow constructed from depth and ego-motion gives the warped and deblurred events in the first column. Columns three and four show disparity at a global (color map shared between rows) and local (color map for only that row) scale. The last two columns show optical flow constructed from depth and ego-motion with global and local color maps, respectively.}
    \label{fig:suppqual8696}
\end{figure*}

\paragraph{Robot experiments.} \cref{fig:suppqual2473,fig:suppqual3233,fig:suppqual4972,fig:suppqual8696} show extended qualitative results for four unseen scenes from flight experiments. Looking at \cref{fig:suppqual2473}, we see that network's ability to maximize the contrast of the image of warped events increases quickly from ``step 0'' (only pre-training) to ``step 3k'' (pre-training + 100 seconds of learning). While this improvement in terms of contrast maximization loss may be mostly due to just learning the correct magnitude of the optical flow (as shown by the change from orange to purple in column 3, and black to red in column 5) through scaling depth and ego-motion, subsequent learning steps improve the depth map in more sophisticated ways, judging from the disappearance of the wrong ``depth gap'' in the center of the disparity images in the third and fourth column. Similar priorities in learning patterns can be seen in the other scenes.

\section{Implementation details}
\label{sec:supptraining}

\paragraph{Training.} For offline training on datasets, we train for 50 epochs with the Adam optimizer and a learning rate of 1e-4. For contrast maximization, we accumulate 10 bins of events, and warp all events to all bin edges. Furthermore, we set the weight for the geometric consistency loss $\lambda=0.05$. For on-device learning, we lower the learning rate to 1e-5. Specifics per dataset are mentioned below. In all cases, event streams are undistorted and rectified.

For MVSEC, we train on \texttt{outdoor\_day2} with input bins of 20~ms of events. We use a batch size of 8, and augment the data with polarity and left-right flips. Training takes around 50 minutes on an RTX 4090.

For DSEC, we train on the daylight sequences in the training set (\texttt{interlaken\_00\_}$^*$ and \texttt{zurich\_city\_\{04,05,06,07,08,11\}\_}$^*$). We leave \texttt{thun\_00\_a} for validation. Because of DSEC's high event density and large frames, we lower the batch size to 4, and bin events to 10~ms frames with a cap of 100k events per bin (if there are more events, we end the bin prematurely; we do not discard events). In addition to polarity and left-right flips, we augment by reversing the time dimension, as we saw that this lessened border artifacts with wrong optical flows. Training takes around 11 hours on an RTX 4090.

For UZH-FPV, we train on the forward indoor sequences (\texttt{indoor\_forward\_\{3,5,6,7,9\}\_davis\_with\_gt} and \texttt{indoor\_forward\_\{8,11,12\}\_davis}). Sequence \texttt{indoor\_forward\_10\_davis\_with\_gt} is left for validation. We use 10~ms bins of events, a batch size of 8 and polarity and left-right flips. Training takes approximately 30 minutes on an RTX 4090.

\paragraph{Network architecture.} We make use of a small convolutional recurrent network to predict depth and ego-motion. The encoder and memory backbone are shared between the depth and ego-motion decoders. \cref{tab:network} lists the details per layer. In addition, we make use of ELU activations as we experienced dying ReLUs. Also, to prevent border artifacts, we use reflect padding for all convolutional layers.

\begin{table*}
\centering
\begin{tabular}{@{}clrrr@{}}
\toprule
                                             & \textbf{Layer type}                                          & \textbf{Input shape}       & \textbf{Output shape}     & \textbf{\# Parameters} \\ \midrule
\multirow{3}{*}{\textbf{Encoder}}            & \texttt{Conv2D(ksize=7, stride=2)}           & $(2, H, W)$       & $(16, H/2, W/2)$ & 1,584         \\
                                             & \texttt{ResidualConv2D(stride=2)}                   & $(16, H/2, W/2)$  & $(32, H/4, W/4)$ & 18,528        \\
                                             & \texttt{ResidualConv2D(stride=2)}                   & $(32, H/4, W/4)$  & $(64, H/8, W/8)$ & 73,920        \\ \midrule
\textbf{Memory}                              & \texttt{ConvGRU}                                    & $(64+64,H/8,W/8)$ & $(64,H/8,W/8)$   & 221,568       \\ \midrule
\multirow{3}{*}{\textbf{Depth}}      & \texttt{Conv2D}                                     & $(64,H/8,W/8)$    & $(64,H/8,W/8)$   & 36,928        \\
                                             & \texttt{Conv2D(bias=False)} w/ \texttt{SoftPlus}     & $(64,H/8,W/8)$    & $(1,H/8,W/8)$    & 576           \\
                                             & \texttt{Upsample(scale=8, "bilinear")} & $(1,H/8,W/8)$     & $(1,H,W)$        &               \\ \cmidrule(l){2-5} 
\multirow{4}{*}{\textbf{Ego-motion}} & \texttt{Conv2D(stride=2)}                           & $(64,H/8,W/8)$    & $(64,H/16,W/16)$ & 36,928        \\
                                             & \texttt{Conv2D(stride=2)}                           & $(64,H/16,W/16)$  & $(64,H/32,W/32)$ & 36,928        \\
                                             & \texttt{Conv2D(bias=False)} w/ \texttt{Identity}    & $(64,H/32,W/32)$  & $(6,H/32,W/32)$  & 3,456         \\
                                             & \texttt{AdaptiveAvgPool2D}                          & $(6,H/32,W/32)$   & $(6,1,1)$        &               \\ \bottomrule
\end{tabular}
\caption{Network layer details. Unless specified otherwise, we use ELU activations, and a kernel size of 3, biases and ``reflect'' padding for convolutional layers. Total parameter count is 430,416.}
\label{tab:network}
\end{table*}

\paragraph{Depth-based control.} We slice depth maps into $K=8$ vertical bins, compute a vector of average inverse depths (disparities) $\bm{d}\in\mathbb{R}^8$, and use it to set a target yaw rate. Furthermore, $\lambda_\text{goal}=0.2$, $\lambda_\text{avoid}=1.0$, $\alpha=0.5$, $\sigma=12.0$.

\paragraph{Drone setup.} We built a 5-inch quadrotor for our robot experiments. The drone is equipped with on-board sensors that provide the flight controller with all relevant information to follow high-level control commands. More specifically, the EKF running on the Kakute H7 Mini flight controller fuses IMU measurements with velocity and height measurements coming from an MTF-01 optical flow/range sensor into a stable position and velocity estimate. This allows a neural network (like our depth network) to give high-level commands like velocity setpoints or rotational rates.

All relevant components on the drone, along with their weight and power consumption, can be found in \cref{tab:hw_components}. Communication on the Orin is handled with ROS2~\cite{macenski2022robot}, which also allows for logging to rosbags, and can be connected via UART to the internal publish-subscribe messaging API of the PX4 flight controller firmware.

The power consumption during flight is measured by keeping track of the total mAh consumed by over two flight tests. Together with the measured power consumption of the Jetson using \texttt{jtop} and the expected maximum power draw of both cameras, this allows us to calculate the power consumption of the drone (see \cref{tab:hw_components}).

\begin{table*}
\centering
\begin{tabular}{@{}llrr@{}}
\toprule
\textbf{Component}     & \textbf{Product} & \textbf {Mass [g]} & \textbf{$\bm{\sim}$Power [W]}               \\ \midrule
Frame                  & Armattan Marmotte 5~inch &  \multirow{7}{*}{455} & \multirow{7}{*}{200$^\dagger$}           \\
Motors                  & Emax Eco II Series 2306 & &        \\
Propellers              & Ethix S5 5~inch  & &               \\
Flight controller      & Holybro Kakute H7 Mini  &   &              \\
Optical flow \& range sensor & MicoAir MTF-01 & & \\
ESC                    & Holybro Tekko32 F4 4in1 mini 50A BL32  &     &   \\
Receiver                    & Radiomaster RP2 V2 ELRS Nano  &     &   \\ \midrule
Battery                & iFlight Fullsend 4S 3000mAh Li-Ion & 208   &  -  \\ \midrule
On-board compute  & NVIDIA Jetson Orin NX 16GB \& DAMIAO v1.1 carrier board & 62 & 9$^*$ \\ \midrule
Event camera     & iniVation DVXplorer Micro  & 22   &      max 0.7$^{\ddagger}$\\ \midrule
Stereo camera	       & Intel RealSense D435i & 75 & max 3.5$^{\ddagger}$  \\ \midrule \midrule 
\textbf{Total} & - & 822 & 213.2 \\ \bottomrule
\end{tabular}
\caption{List of hardware components used during robot experiments. Power consumption estimates are obtained from Jetson's \texttt{jtop}$^*$, battery drain during flight experiments$^\dagger$, or component datasheets$^\ddagger$.}
\label{tab:hw_components}
\end{table*}

\end{document}